\documentclass[runningheads]{llncs}
\usepackage{graphicx}
\usepackage[pdfa]{hyperref}
\hypersetup{
colorlinks=true,
filecolor=blue,
urlcolor=blue
}

\usepackage{amsmath}
\newcommand{\XX}{\mathbf{X}}
\newcommand{\YY}{\mathbf{Y}}
\newcommand{\FF}{\mathbf{F}}
\begin{document}
\title{Lie Transform--based Neural Networks for Dynamics Simulation and Learning}
\titlerunning{Neural Networks for Dynamics Simulation and Learning}
%
\author{Andrei Ivanov\inst{1} \and
Alena Sholokhova \and
Sergei Andrianov\inst{1} \and
Roman Konoplev-Esgenburg}
\authorrunning{A. Ivanov et al.}
%
\institute{Saint Petersburg State University, Saint Petersburg, Russia\\
\email{05x.andrey@gmail.com}}

\maketitle              
\begin{abstract}
In the article, we discuss the architecture of the polynomial neural network that corresponds to the matrix representation of Lie transform. The matrix form of Lie transform is an approximation of the general solution of the nonlinear system of ordinary differential equations. The proposed architecture can be trained with small data sets, extrapolate predictions outside the training data, and provide a possibility for interpretation. We provide a theoretical explanation of the proposed architecture, as well as demonstrate it in several applications. We present the results of modeling and identification for both simple and well-known dynamical systems, and more complicated examples from price dynamics, chemistry, and accelerator physics. From a practical point of view, we describe the training of a Lie transform--based neural network with a small data set containing only 10 data points. We also demonstrate an interpretation of the fitted neural network by converting it to a system of differential equations.

\end{abstract}
\section{Introduction}
Modeling and control of complex dynamical systems require techniques for consideration of nonlinearities and uncertainties. On the face of it, artificial neural networks could provide a suitable approach for learning dynamical systems. The applications 
cover different problems, such as solving ordinary differential equations (ODEs) \cite{ref_1,ref_2}, signal processing \cite{ref_3}, feedback control systems \cite{ref_4}, 
modeling and identification \cite{ref_5}, and others. 
In the article \cite{ref_5a}, the neural network is trained to satisfy the differential operator, initial 
and boundary conditions for the partial differential equation. The backpropagation technique through an ODE solver is proposed in \cite{ref_5b}. 
The comparison of the recent research on solving differential equation with neural networks can be found in \cite{ref_5c}.

In the article, we describe a neural architecture that differs from the described above techniques. Firstly, it is not necessary to train the proposed network for simulation purposes. If the differential equations are provided, the weights of the network can be directly computed. Secondly, we completely avoid numerical ODE solvers in both simulation and data-driven system learning by describing the dynamics with maps instead of numerical step-by-step integrating.


The proposed architecture is a neural network representation of a Lie propagator for dynamical systems integration that is introduced in \cite{ref_6} and is commonly used in the charged particle dynamics simulation.
We consider dynamical systems that can be described by nonlinear ordinary differential equations, 
\begin{equation}
\label{odesystem}
\frac{d}{dt}\XX = \FF(t, \XX) = \sum_{k=0}^{\infty} P^{1k}(t)\XX^{[k]},
\end{equation}
where $t$ is an independent variable, $\XX \in R^n$ is a state vector, and $\XX^{[k]}$ means $k$-th Kronecker power of vector $\XX$. There is an assumption that function $\FF$ can be expanded in Taylor series with respect to the components of $\XX^{[k]}$.

\section{Proposed Neural Network}

\subsection{Matrix Form of Lie Transform}
\label{sec21}

The solution of (\ref{odesystem}) in its convergence region can be presented in the series \cite{ref_6,ref_7b},

\begin{equation}
\label{Lie}
\XX(t|t_0) =  \mathcal M(t|t_0) \circ \XX_0 =  \sum_{k=0}^{\infty} M^{1k}(t|t_0)\XX_0^{[k]},
\end{equation}
where $\XX_0 = \XX(t_0)$ In \cite{ref_7}, it is shown how to calculate matrices $M^{1k}$ by introducing new matrices $P^{ij}$.
The main idea is replacing 
(\ref{odesystem}) by the equation
\begin{equation}
\label{Lie_map}
\frac{d}{dt} M^{ik}(t|t_0) = \sum_{j=i}^{k} P^{ij}(t)M^{jk}(t|t_0),\;1\leq i < k.
\end{equation}
This equation should be solved with initial condition $M^{kk}(t_0) = I^{[k]},\;M^{jk}(t_0) = 0, j\neq k$, where $I$ is the identity matrix. Theoretical estimations of accuracy and convergence of the truncated series in solving of ODEs can be found in \cite{ref_6,ref_7b}. 


The transformation $\mathcal{M}$ can be considered as a discrete approximation of the evolution operator of (\ref{odesystem}) for initial time $t_0$ and interval $\Delta t$. This means that the evolution of the state vector $\XX_0 = \XX(t_0)$ during time $\Delta t$ can be approximately calculated as $\YY = \mathcal{M} \circ \XX_0$. Hence, instead of solving the system of ODEs numerically, one can apply a calculated map and avoid a step-by-step integrating.

\subsection{Neural Network Representation of Matrix Lie Transform}
The proposed neural network implements map $\mathcal M : \XX\rightarrow \YY$ in form of
\begin{equation}
	\label{nnmap}
	\YY = W_0 + W_1\,\XX+W_2\,\XX^{[2]}+\ldots+W_k\,\XX^{[k]},
\end{equation}
where $\XX, \YY \in R^n$, $W_i$ are weight matrices, and $\XX^{[k]}$ means $k$-th the Kronecker power of vector $\XX$. For a given system of ODEs (\ref{odesystem}), one can compute matrices $W_i = M^{1k}$ in accordance with (\ref{Lie_map}) up to the necessary order of nonlinearity. 

\begin{figure}
\centering
\includegraphics[width=0.55\textwidth]{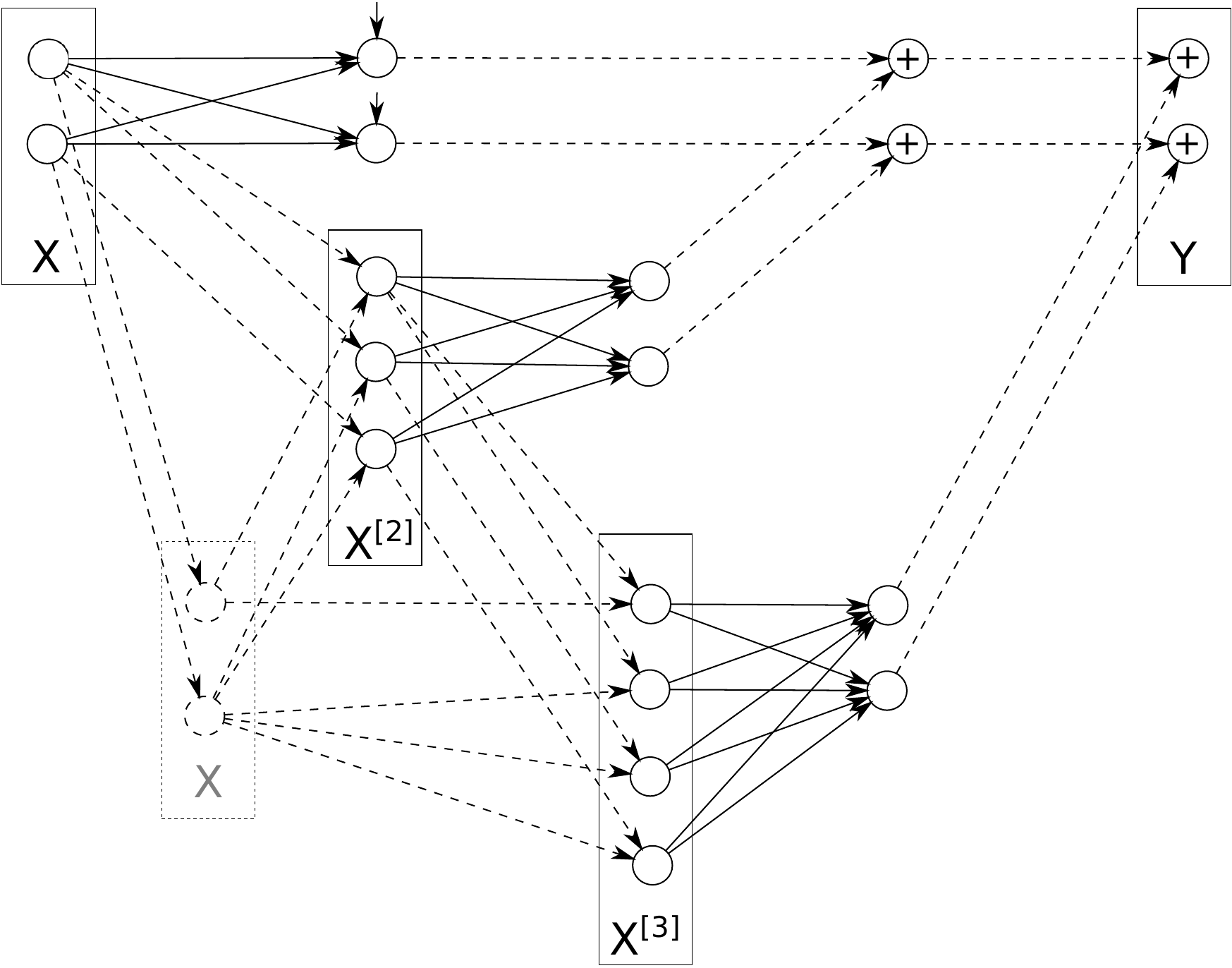}
\caption{Neural network representation of third order matrix Lie map.} \label{fig1}
\end{figure}

Fig.~\ref{fig1} presents a neural network for map (\ref{nnmap}) up to the third order of nonlinearities for a two-dimensional state. In each layer, the input vector $\XX = (x_1, x_2)$ is consequently transformed into $\XX^{[2]} = (x_1^2, x_1x_2, x_2^2)$ and $\XX^{[3]} = (x_1^3, x_1^2x_2, x_1x_2^2$, $x_2^3)$ where weighted sum is applied. The output Y equals to the sum of results from every layer. In the example, we reduce Kronecker powers for decreasing of weights matrices dimension (e.g., $\XX^{[2]} = (x_1^2, x_1x_2, x_2x_1, x_2^2) \rightarrow (x_1^2, x_1x_2, x_2^2)$).

\subsection{Fitting Neural Network}
To fit a proposed neural network, the training data is presented as a multivariate time series (table~\ref{table1}) that describes the evolution of the state vector of the dynamical system in a discrete time. In a general case, each step $t_i \rightarrow t_{i+1}$ should be described as map $\mathcal M_i(t_i): \XX_i \rightarrow \XX_{i+1}$, but if the system (\ref{odesystem}) is time independent, then weights $W_i$ depends only on time interval $\Delta t = t_{i+1} - t_i$.

In the article, we consider only autonomous systems of ODEs with constant discretization $\Delta t =const$. This allows using Lie transformed--based neural network $\mathcal M = \mathcal M(\Delta t)$ with the shared across time weight matrices $W_i$. A  time-dependent right-hand side in (\ref{odesystem}) can be considered by introducing a deeper network architecture.

\begin{table}
\centering
\caption{Discrete states of a dynamical system for training the proposed network.}\label{table1}
\begin{tabular}{lcccr}
\hline
$t_0$ & $t_1$ & $\ldots$ & $t_{m-1}$ & $t_m$ \\
\hline
$x_0(t_0)$ & $x_0(t_1)$ & $\ldots$ & $x_0(t_{m-1})$ & $x_0(t_m)$ \\
$x_1(t_0)$ & $x_1(t_1)$ & $\ldots$ & $x_1(t_{m-1})$& $x_m(t_m)$ \\
$\ldots$   &            & $\ldots$ & &            \\
$x_n(t_0)$ & $x_n(t_1)$ & $\ldots$ & $x_n(t_{m-1})$ & $x_n(t_m)$ \\
\hline
INPUT $\rightarrow$ & $\mathcal M_1$ $\rightarrow$ & $\ldots$ & $\rightarrow$ $\mathcal M_m$ $\rightarrow$ & OUTPUT \\
\hline
\end{tabular}
\end{table}

On these assumptions, the input of the neural network is state $\XX(t_i)$, and the output is $\XX(t_{i+1})$. In all provided examples, the loss function is mean squared error and the Adamax algorithm is used for fitting.
\\

\noindent
\textbf{Supplementary code}\\
\href{https://github.com/andiva/DeepLieNet/blob/master/core/Lie.py}{https://github.com/andiva/DeepLieNet/blob/master/core/Lie.py}\\
\href{https://github.com/andiva/DeepLieNet/blob/master/core/LieMapBuilder.py}{https://github.com/andiva/DeepLieNet/blob/master/core/LieMapBuilder.py}\\


\section{Simulation of Dynamical Systems}
In the section, we describe how matrix Lie map can replace numerical methods for solving well-known systems of ODEs. 
We consciously limit ourselves to a visual comparison of the approaches. The examples of the accuracy estimation can be found in  \cite{ref_5c}.

\subsection{Simple Models}
To demonstrate the application of matrix Lie maps for dynamics simulation, we consider three simple dynamical systems. The Lotka--Volterra system is taken in form of  $x'=-y-xy,\;y'=x+xy$ that can be derived from the classical form by a change of variables. These equations are commonly used for the description of population dynamics in biological, social, and economic systems. The Van der Pol oscillator is defined as $x'=y,\;y'=x'-x-x^2x'$ and can be used for the description of pneumatic hammer, steam engine, periodic occurrence of epidemics, economic crises, depressions, and heartbeat. The Henon--Heiles model is an example of systems where dynamical chaos arises. The system of differential equations can be described as $q'_1=p_1$, $q'_2=p_2$, $p'_1=-q_1-2q_1q_2$, $p'_2=-q_2-q_1^2+q_2^2$. The chaos theory has applications in meteorology, physics, environmental science, computer science, engineering, and philosophy.

\begin{figure}
\centering
\includegraphics[width=0.40\textwidth]{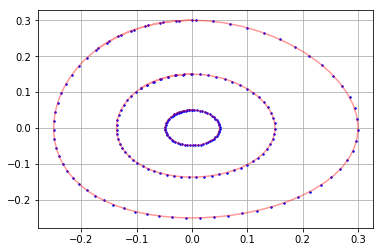}
\includegraphics[width=0.40\textwidth]{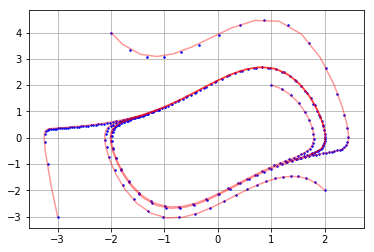}
\caption{Simulation of dynamics in phase space for the Lotka--Volterra equation (left) and the Van der Pol oscillator (right). The red lines correspond to the traditional Runge--Kutta method. The blue dots are for simulation by the matrix Lie map.}
\label{fig2}
\end{figure}

\begin{figure}
\centering
\includegraphics[width=0.80\textwidth]{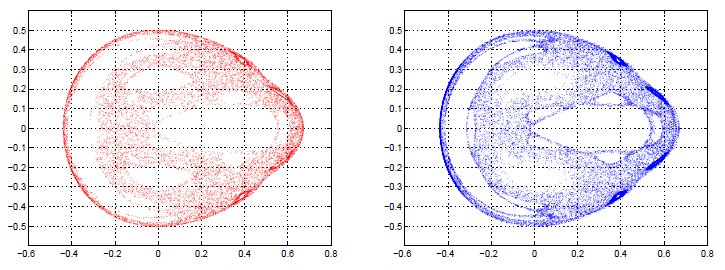}
\caption{Simulation of dynamical chaos by a traditional numerical method (left) and a matrix Lie transform (right).}
\label{fig3}
\end{figure}

\noindent
In Figs. \ref{fig2} and \ref{fig3}, results of the simulation of these systems are shown. Fig.~\ref{fig2} represents the phase space dynamics (in $x(t),y(t)$ coordinates) of the Lotka--Volterra system and the Van der Pol oscillator. Fig. \ref{fig3} shows a Poincaré map ${(q_2, p_2) | q_1 = 0}$, which is calculated by integrating the initial state vector $q_1 = 0.000, q_2 = 0.670, p_1 = 0.093$, and $p_2 = 0.000$. On both figures, the red lines and dots correspond to the numerical integration using Runge--Kutta method of fourth order. The blue dots are for simulation by the matrix Lie transform of third order of nonlinearity.

\subsection{Biochemical Reaction Simulation}
In this example, we demonstrate results of simulation of a biochemical system that is described in \cite{ref_8} and represents the influence of the Raf kinase inhibitor protein (RKIP). In the article, the influence of RKIP is investigated via numerical analysis of nonlinear ordinary differential equations using the MATLAB ode45 function that is based on step-by-step integration. Instead of using a step-by-step numerical integration method, one can build a polynomial neural network and utilize it for system simulation.

\begin{figure}
\centering
\includegraphics[width=0.9\textwidth]{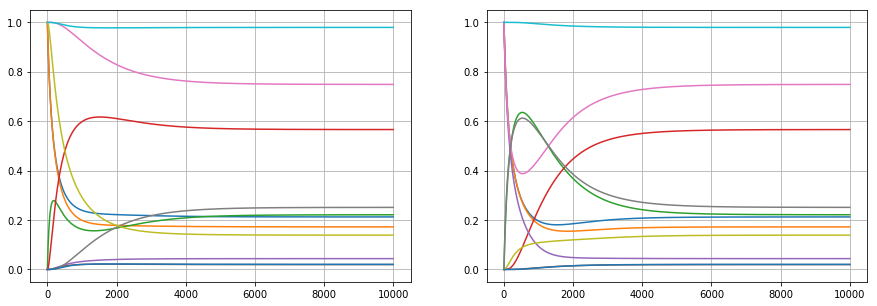}
\caption{Simulation of biochemistry reaction with Lie transform--based neural network.}
\label{fig4}
\end{figure}

The system of differential equation consists of 11 nonlinear equations that describe the biochemical network. We built a second-order Lie map for this system and used it for simulation with the initial condition from \cite{ref_8}. The results of the simulation are shown in Fig.~\ref{fig4} and have a good coincidence with the ones presented in \cite{ref_8}.
\\

\noindent
\textbf{Supplementary code}\\
\href{https://github.com/andiva/DeepLieNet/blob/master/demo/SModels.ipynb}{https://github.com/andiva/DeepLieNet/blob/master/demo/SModels.ipynb}\\

\noindent
\href{https://github.com/andiva/DeepLieNet/blob/master/demo/biochemistry.ipynb}{https://github.com/andiva/DeepLieNet/blob/master/demo/biochemistry.ipynb}\\


\section{Learning Dynamical Systems from Data}
In this section, we describe the examples of the application of Lie transform--based neural networks for data-driven identification of dynamical systems.
\subsection{Epidemic Dynamics}
For this example, we first generate data from the equations of the SIR epidemic model \cite{ref_9}. The model consists of three compartments: $S$ for the number susceptible, $I$ for the number of infectious, and $R$ for the number recovered.
\begin{equation}
\label{SIR}
\frac{dS}{dt} = -\beta\frac{IS}{N},\;\frac{dI}{dt} = \beta\frac{IS}{N}-\gamma I,\;\frac{dR}{dt} = \gamma I.
\end{equation}
We consider a system with parameters $\beta = 5,\gamma = 0.1$, and $N = 10$ on time interval [0; 10]. For the data generation, we use traditional Runge--Kutta methods of fourth order with time step $\Delta t = 0.1$. We define the training set as a particular solution of the system with initial condition $S(0) = 0.99, I(0) = 0.01$, and $R(0) = 0$. The two testing sets were generated from the system as the solutions start with new initial condition $S(0) = 0.4, I(0) = 0.1, R(0) = 0$ for test1, and $S(0) = 1, I(0) = 0, R(0) = 0$ for test 2. After data is generated, we do not use differential equation further.

\begin{figure}
\centering
\begin{minipage}{0.07\textwidth}
\vskip 24pt
\small{train}\\
\vskip 10pt
\small{test 1}\\
\vskip 10pt
\small{test 2}\\
\end{minipage}
\begin{minipage}{0.92\textwidth}
\small{\hskip 30pt LSTM neural network \hskip 45pt Lie transform--based neural network}\\
\vskip 1pt
\hskip 10pt
\includegraphics[width=0.9\textwidth]{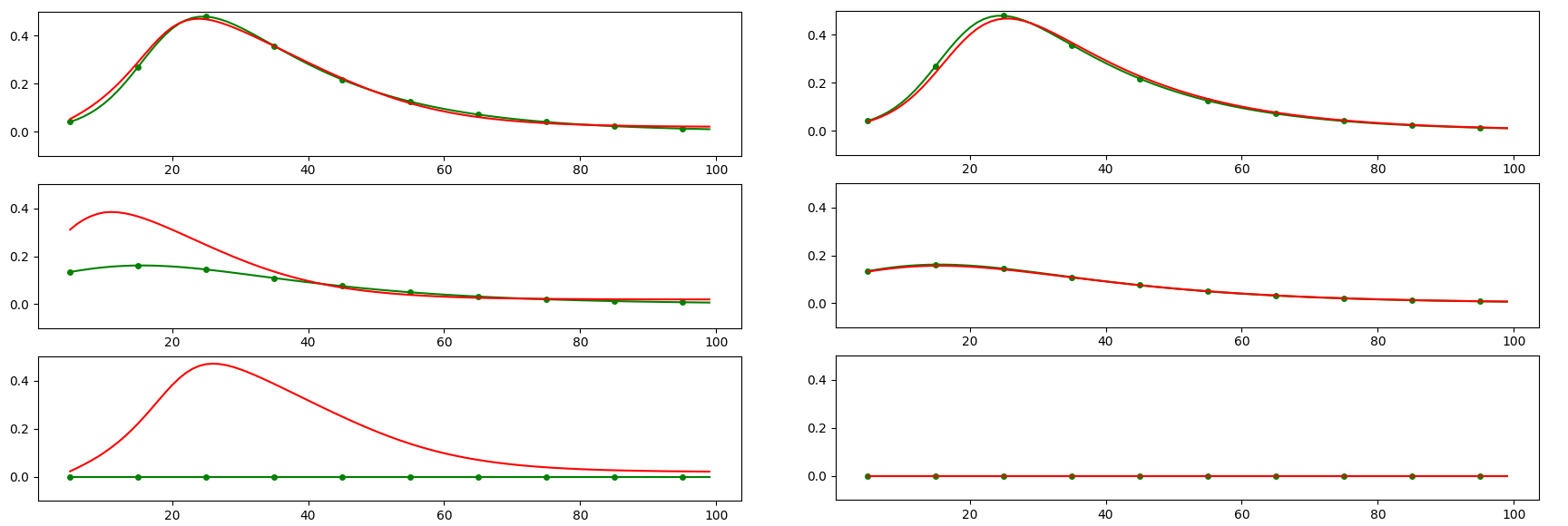}
\end{minipage}

\caption{Memorization of training data by LSTM neural network (left) and generalization of dynamics by Lie transform--based neural network (right). Plots contain the number of infectious (I) generated by equations (green) and predictive models (red).}
\label{fig5}
\end{figure}

To compare the proposed approach with traditional architectures, the LSTM and Lie transform--based neural networks have been fitted only with the training solution. Then the prediction for testing initial conditions that were not presented during fitting is examined. The neural networks configurations can be found following the link provided at the end of the article.

As shown in Fig.~\ref{fig5}, the LSTM neural network just memorized training data. It tends to predict the same solution (training one) regardless of initial conditions, while a Lie transform--based neural network is able to correctly predict the dynamics for previously unseen initial conditions. Moreover, it preserves the physical properties of the system and can recognize the fixed point that corresponds to the absence of the epidemic.

\subsection{iPad and iPhone Sales}
In the article \cite{ref_10} the authors investigate the dynamics of iPhone and iPad sales with differential equations. They suggest analytic formulas for systems of nonlinear ODEs and fit parameters based on time-series data. This is a traditional approach for system identification. On the other hand, using the described above technique one can identify dynamics utilizing Lie transform--based neural network without knowledge of appropriate equations.

\begin{figure}
\centering
\includegraphics[width=0.6\textwidth]{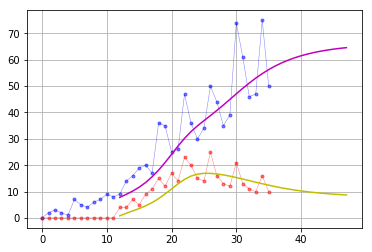}
\caption{Identification of iPhone and iPad sales with Lie transform--based network of the fifth order of nonlinearity (dots for training data, lines for prediction).}
\label{fig6}
\end{figure}
\noindent
In this example, we generated data for iPad and iPhone sales from the plots presented in the article \cite{ref_10}. Then we fitted a fifth order Lie transform--based network with this data. Note that in this case, we did not use any specific assumption on the possible view of equations as made in the original article. The order of nonlinearities is chosen based on the experimentation.

After fitting the neural network, one can receive a Lie map
\begin{equation*}
\begin{aligned}
\begin{pmatrix}
x_{i+1}\\
y_{i+1}\\
\end{pmatrix}
=
\begin{pmatrix}
3.18e-05\\
3.82-03
\end{pmatrix}+
\begin{pmatrix}
0.62&-0.37\\
-0.16&0.45
\end{pmatrix}
\begin{pmatrix}
x_i\\
y_i\\
\end{pmatrix}
+\\
\begin{pmatrix}
-0.36&-0.22&-0.22&-1.25\\
-0.094&-0.27&-0.27&-1.09
\end{pmatrix}
\begin{pmatrix}
x_i^2\\
x_iy_i\\
y_ix_i\\
y_i^2\\
\end{pmatrix}+\ldots,
\end{aligned}
\end{equation*}
where $x_i$ and $y_i$ are sales for iPhone and iPad, respectively, at time $t_i$.
\\

\noindent
\textbf{Supplementary code}\\
\href{https://github.com/andiva/DeepLieNet/blob/master/demo/SIR.ipynb}{https://github.com/andiva/DeepLieNet/blob/master/demo/SIR.ipynb}\\
\href{https://github.com/andiva/DeepLieNet/blob/master/demo/iPhoneiPad.ipynb}{https://github.com/andiva/DeepLieNet/blob/master/demo/iPhoneiPad.ipynb}


\section{Applications}
This section is devoted to the practical applications of the developed technique. Initially, we developed the proposed method for the high-performance simulation of charged particle dynamics. In this article, we briefly mention the key concepts of applying Lie map for modeling of the particle accelerators and storage rings. The second application corresponds to learning a production dynamics with only 10 data points and providing model interpretation. The example is taken from the cosmetic industry.
\subsection{Charged Particle Accelerators}
Charged particle accelerator consists of a number of physical equipment (e.g., quadrupoles, bending magnets, and others
). The design of accelerators and nonlinear dynamics investigation require an accurate computer model of such a complicated system.

\begin{figure}
\centering
\includegraphics[width=0.9\textwidth]{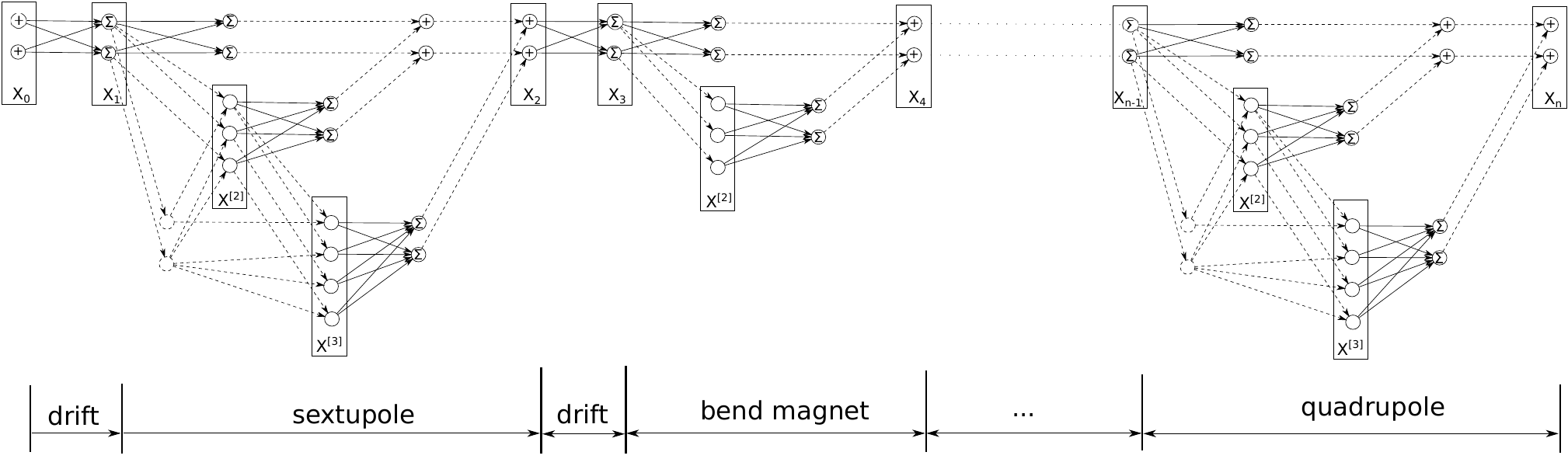}
\caption{ A neural network representation of a charged particle accelerator.}
\label{fig8}
\end{figure}

The particle dynamics in the physical control element can be described by a system of ODEs that has a complex nonlinear form. For instance, the equation of a particle  motion depends on electromagnetic fields and has a 9-dimensional state vector for spin-orbit dynamics.
For long-term dynamics investigation, the traditional step-by-step numerical methods are not suitable because of the performance limitation. Instead of solving differential equations directly, one can estimate a matrix Lie map for each control element in an accelerator. By combining such maps consequently, one can obtain a deep polynomial neural network that represents the whole accelerator ring (see Fig.~\ref{fig8}).

The Lie transform--based mapping approach is commonly used in accelerator physics. For example, in articles \cite{ref_11,ref_12}, the proposed method for simulation of nonlinear spin-orbit dynamics in EDM (electric dipole moment) search project is discussed.

\subsection{Bath Bombs and Bath Fizzies}
Consider a production of a chemical product that implies mixing of 11 source components such as baking soda, olive oil, SLES, water, and others. The known set of component rates in the amount of 10 data points (bath bombs) that leads to the stable characteristic of the product after one minute of mixing is known. Having this information, one needs to produce a new product (bath fizzies) that has a modified SLES component that is equal to zero (see table~\ref{table2}).
\begin{table}
\centering
\caption{Bath bombs (training) and bath fizzies data sets.}\label{table2}
\begin{tabular}{lcccc}
& Bath bombs\\
\hline
 & 1& 2& $\ldots$& 10 \\
\hline
Baking soda & 0.74& 0.83& $\ldots$& 0.65 \\
$\ldots$ &  &  & $\ldots$&   \\
Olive oil & 0.57& 0.77& $\ldots$& 0.60 \\
SLES & 0.76& 0.60& $\ldots$& 0.46 \\
Water & 0.09& 0.14& $\ldots$& 0.08 \\
\hline
\end{tabular}
\hskip 16pt
\begin{tabular}{lcccc}
& Bath fizzies\\
\hline
 & 1& 2& $\ldots$& 10 \\
\hline
Baking soda & 0.74& 0.83& $\ldots$& 0.65 \\
$\ldots$ &  &  & $\ldots$&   \\
Olive oil & 0.57& 0.77& $\ldots$& 0.60 \\
\textbf{SLES} & \textbf{0.00}& \textbf{0.00}& $\ldots$& \textbf{0.00} \\
\textbf{Water} & \textbf{?}& \textbf{?}& $\ldots$& \textbf{?} \\
\hline
\end{tabular}
\end{table}

\noindent
The first challenge in this problem is a limited dataset with only 10 points. The second one is the extrapolation issue. One has to build a model with data that contain SLES and provide predictions where SLES is not presented. This makes it almost impossible to use traditional machine learning methods.

At the same time, a Lie transform--based neural network allows building a model that can provide reasonable results. To achieve this, one has to make two assumptions. Firstly, we consider production as a continuous process that can potentially be described by a system of ODEs. We also assume that these equations are time-independent, allowing us to consider constant Lie map for each time step. Secondly, the initial components in bath bombs and bath fizzies are considered as different initial conditions of the dynamical system.

Under these assumptions, one can represent available data points as a dynamical process from $t_0 = 0$ to $t = 1$ minute with discrete time step $\Delta t = 0.005$. The time series representation of first data point is shown in table~\ref{table3}, where $\XX(0)$ is a state vector of the initial components ($x_0$ is baking soda, $\ldots$, $x_9$ is SLES, $x_{10}$ is water, and $x_{11}$ is product stability), and $\mathcal M$ is a Lie transform-based layers that represent the dynamics. NA means not available and implies hidden process states. Thus, for each of 10 samples, as an input for the neural network, we use the initial component rates. The output is the stability $x_{11}$ of the product.

Because of our assumptions, we can model the production process by a Lie transform--based neural network that consists of 199 consistent Lie maps $\mathcal M$ with shared across time weight matrices (table~\ref{table3}). During fitting of the Lie map, the neural network recovers the dynamics and estimates data points that are marked as NA. After the neural network is fitted with bath bombs, one can use it to predict the dynamics of bath fizzies. The optimal water rates $x_{10}$  with modified SLES component $x_9 = 0$ can be found by varying input variable $x_{10}$ because of the preservation of output variable $x_{11} = 1$ at the final mixing time $t_{199}$.
\begin{table}
\centering
\caption{Time series representation of the first point from training data set.}\label{table3}
\begin{tabular}{lccccc}
& \multicolumn{5}{c}{Bath bombs}\\
\hline
&$t_0$ & $t_1$ & $\ldots$ & $t_{198}$ & $t_{199}$ \\
\hline
$x_0$ & $0.74$ & NA & $\ldots$ & NA& NA \\
$x_1$ & $0.57$ & NA & $\ldots$& NA& NA \\
$\ldots$&&  & $\ldots$           & &            \\
$x_{9}$ & 0.76 & NA & $\ldots$ & NA& NA \\
$x_{10}$ & 0.09 & NA & $\ldots$ & NA& NA \\
$x_{11}$ & 0.00 & NA & $\ldots$ & NA& 1.00 \\
\hline
\hline
\\
& $\XX(0)\rightarrow$ & $\mathcal M\rightarrow$ &  $\ldots$ & $\mathcal M\rightarrow$ & $\XX(1)$
\end{tabular}
\hskip 20pt
\begin{tabular}{lccccc}
& \multicolumn{5}{c}{Bath fizzies}\\
\hline
&$t_0$ & $t_1$ & $\ldots$ & $t_{198}$ & $t_{199}$ \\
\hline
$x_0$ & $0.74$ & NA & $\ldots$ & NA & NA \\
$x_1$ & $0.57$ & NA & $\ldots$& NA & NA \\
$\ldots$&&  & $\ldots$           & &            \\
$x_{9}$ & \textbf{0.00} & NA & $\ldots$ & NA& NA \\
$x_{10}$ & \textbf{?} & NA & $\ldots$ & NA& NA \\
$x_{11}$ & 0.00 & NA & $\ldots$ & NA& 1.00 \\
\hline
\hline
\\
 & $\XX(0)\rightarrow$ & $\mathcal M\rightarrow$ &  $\ldots$ & $\mathcal M\rightarrow$ & $\XX(1)$ \\
\end{tabular}
\end{table}

\noindent
The predictions provided by traditional machine learning method are presented in table~\ref{table4}. Note that the results provided by traditional methods are expected by their design but are physically incorrect. Linear regression provides nonphysical negative rates. Decision tree predicts the same values as in training data. Support vector regression provides almost constant value close to the mean value of water in the training set. Only considering data points as initial conditions of a dynamical process provides physically explainable growth of necessary water in case of SLES absence. This result is also approved by chemical engineers.
\begin{table}
\centering
\caption{Prediction of water for bath fizzies provided by different methods.}\label{table4}
\begin{tabular}{lcccc}
\hline
 & 1& 2& $\ldots$& 10 \\
\hline
Linear Regression & -0.23& -0.12& $\ldots$& -0.13 \\
Decision Tree & 0.76& 0.60& $\ldots$& 0.46 \\
SVR & 0.11& 0.11& $\ldots$& 0.11 \\
\textbf{Lie Map NN} & \textbf{\;\;\;0.64\;\;\;}& \textbf{\;\;\;0.84\;\;\;}& $\ldots$& \textbf{\;\;\;0.36\;\;\;} \\
\hline
\end{tabular}
\end{table}

\noindent
There is also a possibility to translate found weight matrices of the neural network to the equations. To implement this, one has to find such formulas for the system of ODEs that provide the found weights after implementing the described in section \ref{sec21} algorithm. For instance, we parameterized the right-hand side of ODEs up to the second order of nonlinearity and derive the following system,
\begin{equation*}
\begin{aligned}
x_0' &= 0,\;\ldots,\;x_{10}' = -a_5x_7x_{11}\\
x_{11}' &= (a_1x_3+a_2x_4+a_3x_5+a_4x_6+a_5x_7+a_6x_8)x_{11} - a_7x_{11},
\end{aligned}
\end{equation*}
which consists of polynomial right-hand sides with 30 parameters $a_i$. This system of ODEs approximately equivalent to the fitted neural network. So it can be used as an interpretation of the neural network. For instance, one can state that baking soda $x_{0}$ is just a parameter, water $x_{10}$ decreases in time during mixing with the velocity that is proportional to other components. While the stability rate has more complex dynamics and depends on more components.
\\

\noindent
\textbf{Supplementary code}\\
\href{https://github.com/andiva/DeepLieNet/blob/master/demo/accelerator.ipynb}{https://github.com/andiva/DeepLieNet/blob/master/demo/accelerator.ipynb}\\
\href{https://github.com/andiva/DeepLieNet/blob/master/demo/BathBombs.ipynb}{https://github.com/andiva/DeepLieNet/blob/master/demo/BathBombs.ipynb}\\
The data points for bath bombs are presented in dimensionless view, as well as the resulting system of ODEs is not provided due to the data protection policy.

\section{Conclusion}

In the article, we demonstrate a general concept of building a neural network representation of dynamical systems. Although we considered ODEs only with polynomial right-hand side, such nonlinear systems are widely used in different fields of automated control, robotics, mechanical systems, biology, chemical reactions, drug development, molecular dynamics.

The greatest advantage of the proposed Lie transform--based neural network is its equivalence to the differential equations. The weight matrices of a neural network correspond to the certain order of nonlinearities in the real system and have a physical explanation.


As soon as the proposed neural architecture has a good coincidence with traditional modeling methods, it can be useful for the investigation of dynamics in unknown parameter space. The promising properties of the proposed technique are the ability to learn dynamics with small training data sets and to interpret the data-driven model by translating it to the system of ODEs.

The questions of noisy data, truncation of matrix Lie transform, accuracy and convergence for larger systems are not discussed in the article. We also do not consider the optimal selection of loss functions and optimization methods for training. These questions should be investigated in further research.

The proposed Lie transform--based neural network, an algorithm for calculating of the map for the system of ODEs, as well as examples, are implemented in Python (Keras/TensorFlow) and presented at GitHub repository: \href{https://github.com/andiva/DeepLieNet}{https://github.com/andiva/DeepLieNet}. 
 
\section{Acknowledgments}
The authors gratefully thank Prof. Dr. Yurij Senichev for his support in the development of Lie transform--based mapping methods for accelerator physics. Also, we appreciate the efforts of Roman Konoplev-Esgenburg for the explanation of the production process of bath bombs and for providing data.
%
%
%

\begin{thebibliography}{12}
\bibitem{ref_1}Lagaris, I.E., Likas, A., Fotiadis, D.I.: Artificial neural networks for solving ordinary and partial differential equations. Tech. rep. (1997),
\href{https://arxiv.org/pdf/physics/9705023.pdf}{https://arxiv.org/pdf/physics/9705023.pdf}, last accessed 2019/03/03.
\bibitem{ref_2}Mall, S., Chakraverty, S.: Comparison of artificial neural network architecture in solving ordinary differential equations. Advances in Artif. Neural Syst. (2013).
\bibitem{ref_3}Lapedes, A., Farber, R.: Nonlinear signal processing using neural networks: Prediction and system modelling. In: Langley, P. (ed.) IEEE International Conference on Neural Networks, San Diego, CA, USA (1987).
\bibitem{ref_4}Lewis, F.L., Ge, S.S.: Neural networks in feedback control systems. Mechanical Engineers' Handbook: Instrumentation, Systems, Controls, and MEMS, vol. 2. John Wiley and Sons, Hoboken, NJ, USA, third edition (2005).
\bibitem{ref_5}Chen, S., Billings, S.A.: Neural networks for nonlinear dynamic system modelling and identification. International Journal of Control 56(2), 319--346 (1992).
\bibitem{ref_5a}Sirignano, J., Spiliopoulos, K.: DGM: A deep learning algorithm for solving partial differential
equations. Journal of Computational Physics (2018).
\bibitem{ref_5b}Chen, R., Rubanova, Y., Bettencourt, J., Duvenaud, D.: Neural ordinary differential equations, \href{https://arxiv.org/pdf/1806.07366.pdf}{https://arxiv.org/pdf/1806.07366.pdf}, last accessed 2019/03/03.

\bibitem{ref_5c} Ivanov, A., Andrianov, S.: Matrix Lie maps and polynomial neural networks for solving differential equations. Submitted to this Proceedings (2019).

\bibitem{ref_6}Dragt, A.: Lie methods for nonlinear dynamics with applications to accelerator physics (2011), \href{http://inspirehep.net/record/955313/files/TOC28Nov2011.pdf}{inspirehep.net/record/955313/files/TOC28Nov2011.pdf}, last accessed 2019/03/03.
\bibitem{ref_7}Andrianov, S.: A role of symbolic computations in beam physics. Computer Algebra in Sc. Comp., Lecture Notes in Computer Science, 6244, 19--30 (2010).
\bibitem{ref_7b}Andrianov, S.: The convergence and accuracy of the matrix formalism approximation. In: Proceedings of ICAP2012, Rostock, Germany, 93--95 (2012).
\bibitem{ref_8}Gilbert, D., Heiner, M.: From Petri nets to differential equations - an integrative approach for biochemical network analysis. Lecture Notes in Computer Science, 181--200 (2006).
\bibitem{ref_9}Martcheva, M.: An introduction to mathematical epidemiology, \href{https://www.springer.com/gb/book/9781489976116}{https://www.springer.com/gb/book/9781489976116}, last accessed 2019/03/03.
\bibitem{ref_10}Guidolin, M., Guseo, R.: On product cannibalization: a new Lotka-Volterra model for asymmetric competition in the ICTs. Tech. rep. (2016), \href{http://paduaresearch.cab.unipd.it/9738/1/GUIDOLINGUSEO\_SETT2016.pdf}{http://paduaresearch.cab.unipd.it/9738/1/GUIDOLINGUSEO\_SETT2016.pdf}, last accessed 2019/03/03.
\bibitem{ref_11}Senichev, Y., Lehrach, A., Maier, R., Zyuzin, D., Berz, M., Makino, K., Andrianov, S., Ivanov, A.: Storage ring EDM simulation: methods and results. In: Proceedings of ICAP2012, Rostock, Germany, 99--103 (2012).
\bibitem{ref_12}Ivanov, A., Andrianov, S., Senichev, Y.: Simulation of Spin-orbit Dynamics in Storage Rings. Journal of Physics: Conference Series, Volume 747, N.~1 (2016).








\end{thebibliography}
%

\end{document}